\documentclass[runningheads]{llncs}

\usepackage{eccv}

\usepackage{eccvabbrv}

\usepackage{graphicx}
\usepackage{booktabs}

\usepackage{multirow}

\usepackage[accsupp]{axessibility}  

\usepackage[pagebackref,breaklinks,colorlinks,citecolor=eccvblue]{hyperref}

\usepackage{orcidlink}

\newcommand{\ours}[0]{DETRAM\xspace}
\newcommand{\oursfull}[0]{End-to-end \textbf{DE}tection, \textbf{T}racking and \textbf{R}ecovery of Hum\textbf{A}n \textbf{M}eshes\xspace}

\usepackage{suffix}
\newcommand\parahead[1]{\vspace{1.5mm}\noindent\textbf{#1.}\medspace}
\WithSuffix\newcommand\parahead*[1]{\vspace{2mm}\noindent\textbf{#1}\medspace}

\usepackage[dvipsnames]{xcolor}

\usepackage{array}
\newcolumntype{L}[1]{>{\raggedright\let\newline\\\arraybackslash\hspace{0pt}}m{#1}}
\newcolumntype{C}[1]{>{\centering\let\newline\\\arraybackslash\hspace{0pt}}m{#1}}
\newcolumntype{R}[1]{>{\raggedleft\let\newline\\\arraybackslash\hspace{0pt}}m{#1}}

\usepackage{wrapfig}

\begin{document}

\title{DETRAM: End-to-end \underline{DE}tection, \underline{T}racking and \underline{R}ecovery of Hum\underline{A}n \underline{M}eshes} 

\titlerunning{DETRAM}

\author{%
Chunggi Lee\inst{1} \and
Seonwook Park\inst{2} \and
Wanhua Li\inst{3} \and
Umar Iqbal\inst{2}\textsuperscript{,*} \and
Hanspeter Pfister\inst{1}\textsuperscript{,*}}


\authorrunning{C.~Lee et al.}

\institute{
\textsuperscript{1}Harvard University \quad
\textsuperscript{2}NVIDIA \quad
\textsuperscript{3}Nanyang Technological University
}


\maketitle

\begingroup
\renewcommand{\thefootnote}{*}
\footnotetext{Corresponding authors.}
\endgroup
\begin{center}
    \includegraphics[width=1.0\textwidth]{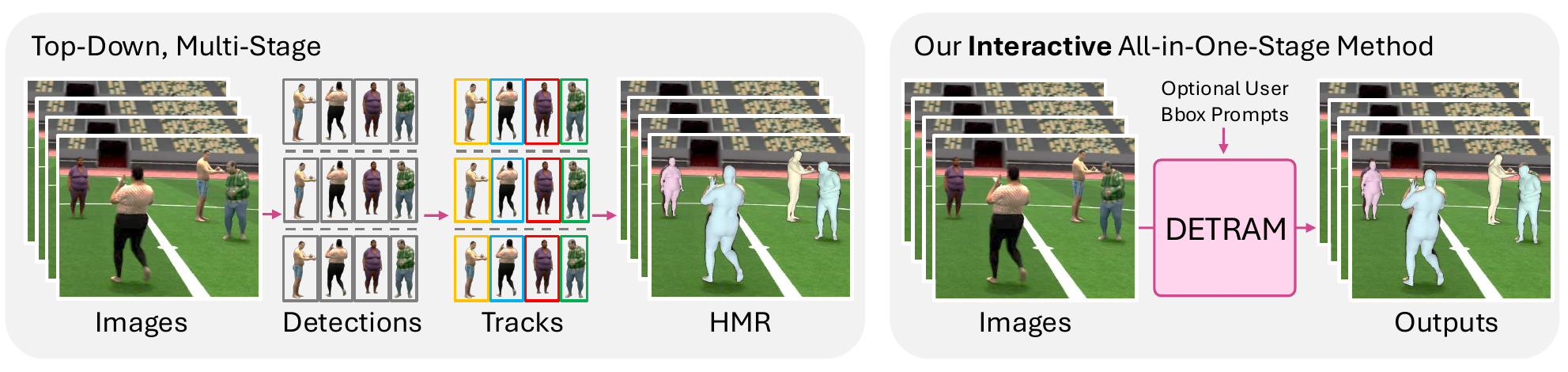}
    \captionof{figure}{Comparison between traditional multi-stage pipelines and our unified one-stage DETRAM framework. Conventional approaches decompose the problem into sequential detection, tracking, and human mesh recovery (HMR) modules, causing error accumulation, increased latency, and difficulty maintaining consistent identities over time. In contrast, DETRAM performs all three tasks jointly in a single feed-forward pass. This enables efficient, end-to-end multi-person reconstruction and  tracking 
    without requiring either an external detector or a separate learned tracking module.
    DETRAM discovers and tracks new people as they enter the scene, while also supporting optional user-prompted tracking, allowing individuals to be added or specified when desired. 
    \label{fig:teaser}
    }
    \vspace{3mm}
\end{center}

\begin{abstract}
\vspace*{-8mm}

In the task of human mesh recovery (HMR), multi-person scenes are particularly difficult to handle due to the many entities that appear and occlusions between them over time.
In particular for video inputs, there is a need to track each entity reliably and consistently. Existing methods rely on pretrained human detection modules, increasing their runtime and limiting the number of tracked entities.
We present \ours, a unified framework for multi-person HMR and tracking that simultaneously detects, reconstructs, and tracks humans across time, both automatically and via user prompts.
\ours uses a single transformer decoder with an identity-consistent set of learnable query embeddings that persist across frames: detection queries discover new people, tracking queries maintain pose and shape for existing individuals, and prompt queries follow user-specified identities.
Our approach achieves state-of-the-art tracking results on PoseTrack21, 3DPW, BEDLAM, and MuPoTS-3D, and competitive reconstruction accuracy on BEDLAM and 3DPW, while uniquely supporting prompt-based tracking of individuals in multi-person scenes. To our knowledge, this is the first method to unify promptability and multi-person HMR with tracking in an end-to-end trainable framework, enabling user-directed human analysis in videos.

\end{abstract}    
\section{Introduction}
\label{sec:intro}

Videos of public places, sporting events, and urban environments typically contain many humans and their movements over time.
Tracking each human and recovering their 3D pose and shape in such crowded videos enables applications in sports analytics, mixed reality, safety monitoring, and human behavior understanding.
However, these scenes are highly challenging: people frequently occlude one another, body parts are truncated by crops or the camera view, and individuals can enter or leave the scene at any time.
These factors often lead to tracking failures or inconsistent 3D reconstructions, especially in long sequences.

Existing methods for multi-person human mesh recovery (HMR) and tracking are mostly built as multi-stage pipelines.
A detector first outputs bounding boxes, which are subsequently cropped and fed into pose or mesh recovery networks. Tracking IDs are then assigned in a separate association stage based on appearance or pose features~\cite{goel2023humans,rajasegaran2021tracking,rajasegaran2022tracking}.
While effective in some settings, this design has several weaknesses.
Bounding box errors or misaligned crops can remove important spatial context, leading to missing limbs or ambiguous associations in crowded scenes.
Moreover, errors propagate across stages, each module may need to be trained independently, and the overall runtime increases.

Recent transformer-based approaches address some of these issues by operating directly on the full image and regressing human pose and shape without cropping, within detection-transformer architectures for multi-person HMR~\cite{baradel2024multi,sun2024aios,su2025sat}.
By reasoning over all humans jointly via attention, these models better preserve inter-human relationships and global context, and they can deliver more accurate multi-person reconstructions in a single frame.
However, they still treat frames independently, with no explicit mechanism for maintaining identity over time or interacting with user-specified individuals.
Extending them to video, therefore, requires an external tracking module or post-hoc association strategy, reintroducing the multi-stage issues described above and making it non-trivial to deploy them directly in streaming video settings.

In this work, we take a unified approach to multi-person HMR and tracking in videos by proposing an ``\oursfull'' Transformer (\ours) architecture. 
Instead of relying on external detectors or separately trained tracking modules, we define an identity-consistent set of learnable query embeddings that persist across time.
Within a transformer-based decoder in line with recent detector-transformer designs~\cite{li2022dn,liu2022dab}, these queries take on three complementary roles:
(1) \emph{detection queries} discover new humans entering the scene;
(2) \emph{tracking queries} carry pose, shape, and identity information forward across frames; and
(3) \emph{prompt queries} are initialized from user-specified spatial prompts (i.e. bounding boxes) and enforce controllable tracking of particular individuals throughout a sequence.
By jointly decoding all query types in a single feed-forward pass per frame, our model simultaneously detects, reconstructs, and tracks all relevant humans without any explicit cropping or any separately learned tracking stage (as illustrated in \autoref{fig:teaser}).

\ours's unified formulation improves robustness to occlusions and crowded scenes, and substantially reduces inference latency compared to multi-stage pipelines.
Concurrent query-based methods for multi-person mesh recovery~\cite{liu2025motions,newell2025comotion} also aim to do end-to-end tracking and reconstruction, but either use clip-level motion queries with a separate learnable temporal module~\cite{liu2025motions} or decouple per-frame detection and pose updates into distinct components~\cite{newell2025comotion}. In contrast, we use a single framewise decoder in which detection, tracking, and prompt queries are decoded jointly, and only our formulation exposes user-specified prompt queries for controllable tracking.
Our approach attains state-of-the-art tracking performance on PoseTrack21, MuPoTS-3D, BEDLAM, and 3DPW, and reconstruction accuracy competitive with or exceeding the best prior methods on BEDLAM and 3DPW, and, to our knowledge, is the first to unify promptability with multi-person HMR and tracking in an end-to-end trainable framework. 

In summary, our contributions are:
(a) we introduce \ours, a single-stage architecture that detects, tracks, and recovers 3D meshes of multiple humans from video, 
(b) this is enabled by a query design involving identity-consistent tracking queries propagated across frames and prompt queries initialized from user inputs; and
(c) experimental results on the PoseTrack21, MuPoTS-3D, 3DPW and BEDLAM datasets show competitive tracking and reconstruction performance.
\section{Related Work}

\parahead{Human mesh recovery}
Early HMR works from monocular images focused on single-person setups and extended to multiple people using scene-level constraints or multi-stage pipelines~\cite{zanfir2018monocular,jiang2020coherent,lin2021end,sun2022putting,hmrKanazawa18,kocabas2021pare,rogez2019lcr,li2020hybrik,choutas2020expose,humanMotionKanazawa19,kolotouros2019spin,kolotouros2019convolutional}. Typical systems run a detector, crop each person, and estimate pose and shape independently.
More recent approaches use one-stage, transformer-based models that operate on full images and regress meshes for all visible people jointly~\cite{sun2024aios,baradel2024multi,su2025sat,Sun_2021_ICCV,sun2022putting,wang2025prompthmr}, which improves robustness in crowded scenes by keeping global context and modeling inter-person interactions.
These methods, however, are still single-frame: they do not maintain identities over time and they offer no mechanism for selecting which individuals to follow in a video.
Our work follows the one-stage, full-frame formulation but treats detection, reconstruction, and identity as a single video-level problem by making a subset of queries persistent across frames.

\parahead{Tracking and reconstructing humans from video}
Several methods combine tracking and 3D reconstruction by lifting 2D detections into a 3D representation and tracking in that space. T3DP~\cite{rajasegaran2021tracking}, PHALP~\cite{rajasegaran2022tracking}, and 4DHumans~\cite{goel2023humans} estimate 3D pose and position per person and then associate tracks in this feature space, while keeping detection, HMR, and tracking as separate modules. More recent frameworks move towards end-to-end multi-person motion capture. Motions-as-Queries (MaQ)~\cite{liu2025motions} uses temporal motion queries over clips and predicts holistic whole-body motion directly from video in an offline manner. CoMotion~\cite{newell2025comotion} combines a strong per-frame detector with a learned pose-update network that updates 3D poses online in crowded scenes. These concurrent methods also adopt query or update-based formulations for multi-person mesh recovery, but do not allow user-specified prompts. Our method has a similar goal, multi-person 3D reconstruction with stably tracked identities, but uses a single DETR-style decoder with explicit detection, tracking, and prompt queries, enabling user-directed tracking.
The concurrent work, SAM 3~\cite{carion2025sam}, also supports promptable video segmentation and tracking, but focuses on category or exemplar-level 2D mask propagation. In contrast, DETRAM detects, reconstructs, and tracks 3D SMPL parameters via a unified decoder, by treating prompts as identity anchors.

\parahead{DETR-style set prediction and query design}
DETR and follow-up work cast perception as set prediction with learned queries and bipartite matching~\cite{carion2020end,liu2022dab,li2022dn}. Variants modify query parameterization (e.g., dynamic anchor-box queries in DAB-DETR~\cite{liu2022dab}) and training (e.g., denoising in DN-DETR~\cite{li2022dn}) to stabilize and speed up learning. Recent multi-person HMR models adopt DETR-style decoders to regress 3D meshes from single images~\cite{sun2024aios,baradel2024multi,su2025sat}, but treat all queries as frame-local. We build on the same set-prediction method, but extend it temporally and structurally: a subset of queries is kept identity-consistent across frames, and explicit prompt queries are added for user control. All query types are optimized jointly with a denoising-style objective, yielding an end-to-end architecture that handles detection, tracking, and mesh recovery of multiple humans in video within a single decoder.
\section{Method}

\begin{figure}[t]  
    \centering
    \includegraphics[width=1.0\linewidth]{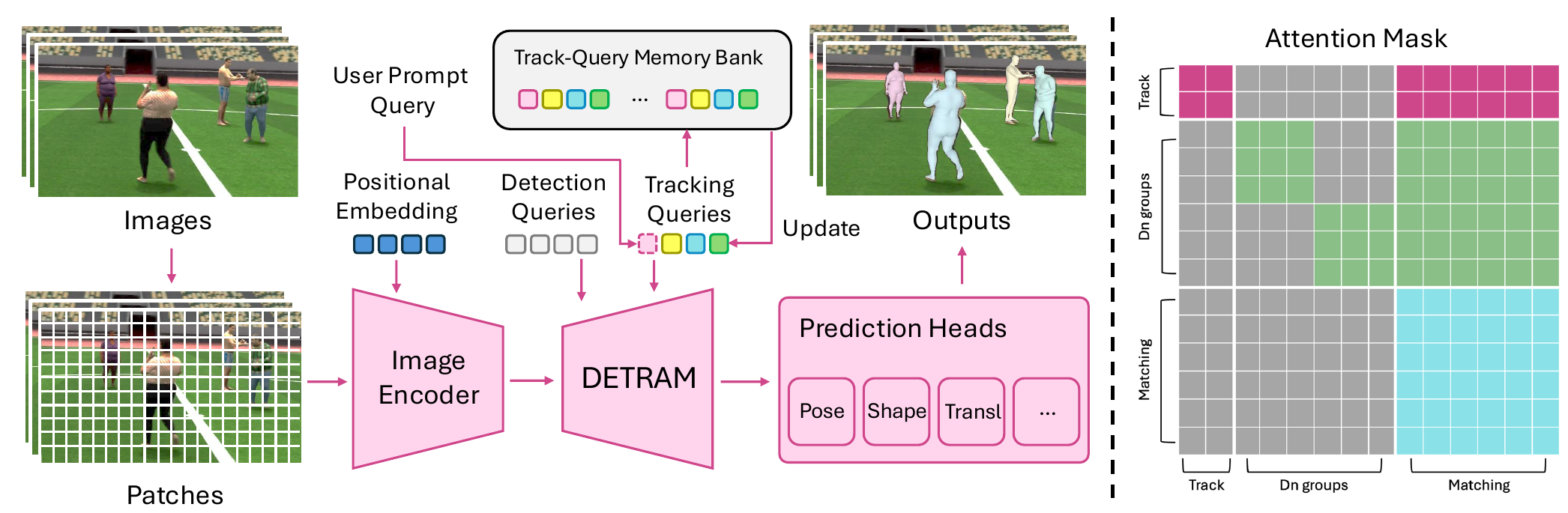}
    \vskip -2mm
    \caption{
    \textbf{Overview of the DETRAM framework.}
Our model (left) encodes input images into visual tokens and jointly processes detection queries (discovering new individuals), tracking queries (identity-persistent embeddings propagated across frames), and optional prompt queries, continuously updated via a track-query memory bank for long-range temporal consistency. Per-person predictions for pose, shape, translation, and confidence are produced through multi-headed regressors. 
The attention-masking scheme (right) separates Hungarian-matching, denoising, and tracking/prompt queries to prevent denoising GT leakage, allowing \ours to learn detection, reconstruction, and identity tracking in a single end-to-end pass while preserving the distinct role of each query type.
    }
    \label{fig:overview}
    \vskip -4mm
\end{figure}

\paragraph{Preliminaries.}
Given an input image sequence $\{I^1, I^2, \dots, I^T\} \in \mathbb{R}^{T\times h\times w\times 3}$ containing an unknown number of people $N_p$, our goal is to recover articulated 3D body meshes with temporally consistent identities. We adopt the SMPL~\cite{loper2015smpl} parametric model to represent each human with pose, shape, and global translation in the camera coordinate system. For each person $i$, the mesh with $N_v=6890$ vertices is defined as, $H_i = \mathrm{SMPL}(\theta_i, \beta_i, \tau_i)$,
where $\theta_i \in \mathbb{R}^{J \times 3}$ encodes the $J=24$ joint rotations (including global orientation), $\beta_i \in \mathbb{R}^{10}$ denotes the low-dimensional shape parameters, and $\tau_i \in \mathbb{R}^{3}$ is the 3D translation in the camera frame.

\parahead{Overview}
We build upon recent DETR-style architectures~\cite{carion2020end, cheng2022masked}, extending them to multi-person human mesh reconstruction and identity tracking. As illustrated in \autoref{fig:overview}, our framework is comprised of an image encoder, a \ours decoder, and multiple prediction heads responsible for estimating confidence scores, bounding boxes, and SMPL parameters.

For each frame, the image is divided into fixed-size patches of dimension $P \times P$. Each patch is embedded into a feature token using patch embeddings and positional encodings, producing a token set $\mathcal{T} = \{t_1, t_2, \dots, t_k\}$, where $k = (h/P) \times (w/P)$. These tokens are processed by a transformer encoder to capture spatial structure across the image. The decoder then operates over a set of learnable queries $\mathcal{Q} = \{q_1, q_2, \dots, q_{N_q}\}$, with $N_q$ selected to accommodate the maximum number of people expected in a scene. Through cross-attention with the encoded features, each query progressively evolves into a latent representation corresponding to an individual human instance.

Decoder outputs are passed through lightweight MLP-based prediction heads, including a confidence head $\mathcal{H}_c$, pose head $\mathcal{H}_p$, shape head $\mathcal{H}_s$, translation head $\mathcal{H}_t$, and bounding box head $\mathcal{H}_b$. These heads jointly regress SMPL parameters (pose $\theta$, shape $\beta$, translation $t$) along with 2D bounding boxes. A separate linear projection predicts per-query confidence scores, and only predictions with confidence above a threshold $\alpha_d$ are retained.



\subsection{Detection and Tracking Queries.}

We tokenize each input image using a DINOv2-based ViT encoder~\cite{oquab2023dinov2}: $F = \text{Encoder}(I)$ with an input resolution of $1288 \times 1288$, producing dense visual tokens that serve as the basis for all subsequent query reasoning.

Our formulation unifies detection and tracking directly within the transformer decoding process through three complementary query types: \emph{detection queries}, \emph{tracking queries}, and \emph{prompt queries}.  
Together, these enable our model to simultaneously discover new people, maintain persistent identities, and integrate user-directed prompts in a single end-to-end framework. 
All three query types share an anchor-based representation: learned for detection, propagated for tracking, and user-provided for prompts.
Each query is represented by an anchor $A_q$, a content embedding $C_q$, and an anchor-derived positional embedding.
Detection queries are learned, while tracking and prompt queries use propagated or user-provided anchors with TMCA-refreshed contents.

\parahead{Detection Queries}
Detection queries discover and localize human instances in each frame.  
Instead of relying on preset anchor shapes or external detectors, we \emph{learn} a set of query anchors.  
Each detection query is parameterized by an anchor box as $A_q = (x_q, y_q, w_q, h_q)$,
where $(x_q, y_q)$ denote the normalized center coordinates and $(w_q, h_q)$ the normalized width and height.  
Each anchor $A_q$ is paired with a content embedding $C_q \in \mathbb{R}^D$ and a positional embedding $P_q \in \mathbb{R}^D$.  
Through cross-attention with visual tokens, these detection queries identify potential human instances and predict their 2D locations, 3D pose, shape, and confidence, acting as the entry point for newly appearing people.  



\parahead{Tracking and Prompt Queries}
%
In dynamic scenes: as people move, enter or exit the frame, or undergo significant appearance changes due to pose variation, the correspondence between detection queries and human entities drifts, leading to fragmented trajectories and identity errors.
To address this, we design our tracking queries to be \emph{identity-persistent}.
At every new frame, each tracking query attends to the detection queries and updates its representation through \emph{Track-Modulated Cross-Attention}.  
This enables the query to absorb relevant spatial context while preserving its identity embedding, ensuring robust temporal consistency even under occlusions or appearance changes.  
The same mechanism naturally extends to user-driven \emph{prompt queries}, allowing explicit control over which individuals to follow throughout the sequence.

\parahead{Track-Modulated Cross-Attention}
To maintain identity consistency across frames, each tracking query updates its content embedding by attending to the detection queries.  Here, $N_t$ denotes the number of tracking queries and $N_d$ denotes the number of detection queries.
Given the set of tracking-query anchors from the previous frame $A^{\text{trk}} = \{A_q^{\text{trk}}\}_{q=1}^{N_t}$ and the detection-query anchors $A^{\text{det}} = \{A_k^{\text{det}}\}_{k=1}^{N_d}$, we compute cross-attention weights as
\vspace{-2pt}
\begin{equation*}
w_{qk} = \mathrm{Softmax}_k\!\left( \phi(A_q^{\text{trk}})^\top \psi(A_k^{\text{det}}) \right),
\label{eq:trk_weight}
\end{equation*}
where $\phi(\cdot)$ and $\psi(\cdot)$ project anchor boxes into a shared embedding space.  
The resulting weights, computed with a top-$k$ softmax for stability, encode both spatial proximity and semantic similarity, allowing each tracking query to selectively aggregate information from the most relevant detection queries.
Each tracking query then updates its content embedding by aggregating information from the detection queries using the computed attention weights:
\vspace{-3pt}
\begin{equation*}
C_q^{\text{trk}} =
\sum_{k \in \mathcal{N}_k(q)} w_{qk} C_k^{\text{det}},
\label{eq:trk_content}
\end{equation*}
where $C_k^{\text{det}}$ denotes the content embedding of the $k$-th detection query, and $\mathcal{N}_k(q)$ denotes the indices of top-$k$ detection queries selected for tracking query $q$.
This update mechanism enables each tracking query to draw identity-relevant cues from the most compatible detections in the current frame, allowing it to adapt to changes in position, appearance, and pose.  


\parahead{Generating Tracking Queries at Training Time}
During training, we construct tracking queries directly from ground-truth annotations, following a strategy inspired by DN-DETR~\cite{li2022dn}.  
Rather than generating multiple denoised variants per instance, we create a single tracking query for each ground-truth person, which serves as the identity anchor for that individual during training.

To improve robustness to localization errors, we perturb the ground-truth bounding boxes by adding Gaussian noise to their center coordinates $(x, y)$ and size parameters $(w, h)$ before converting them into tracking-query anchors.  
We further apply stochastic supervision by sampling only a subset of ground-truth instances at each iteration, instead of supplying tracking queries for all annotated instances. 
As a result, our model becomes robust to imperfect or noisy bounding boxes from previous frames.

\parahead{User Prompting}
At inference time, we optionally allow user-provided box prompts to initialize tracking-query anchors.
A user may provide a bounding box (e.g., in the first frame or after a missed track) to initialize or re-initialize a tracking query for the target person, enabling the model to maintain or recover that identity in the remaining frames. DETRAM currently supports bounding-box prompts. Other prompting modalities such as masks, points, or text can be incorporated (when supported) via external mapping or grounding methods.

\subsection{DETRAM}

\begin{wrapfigure}{r}{0.42\textwidth}
    \centering
    \vspace*{-10mm}
    \includegraphics[width=\linewidth]{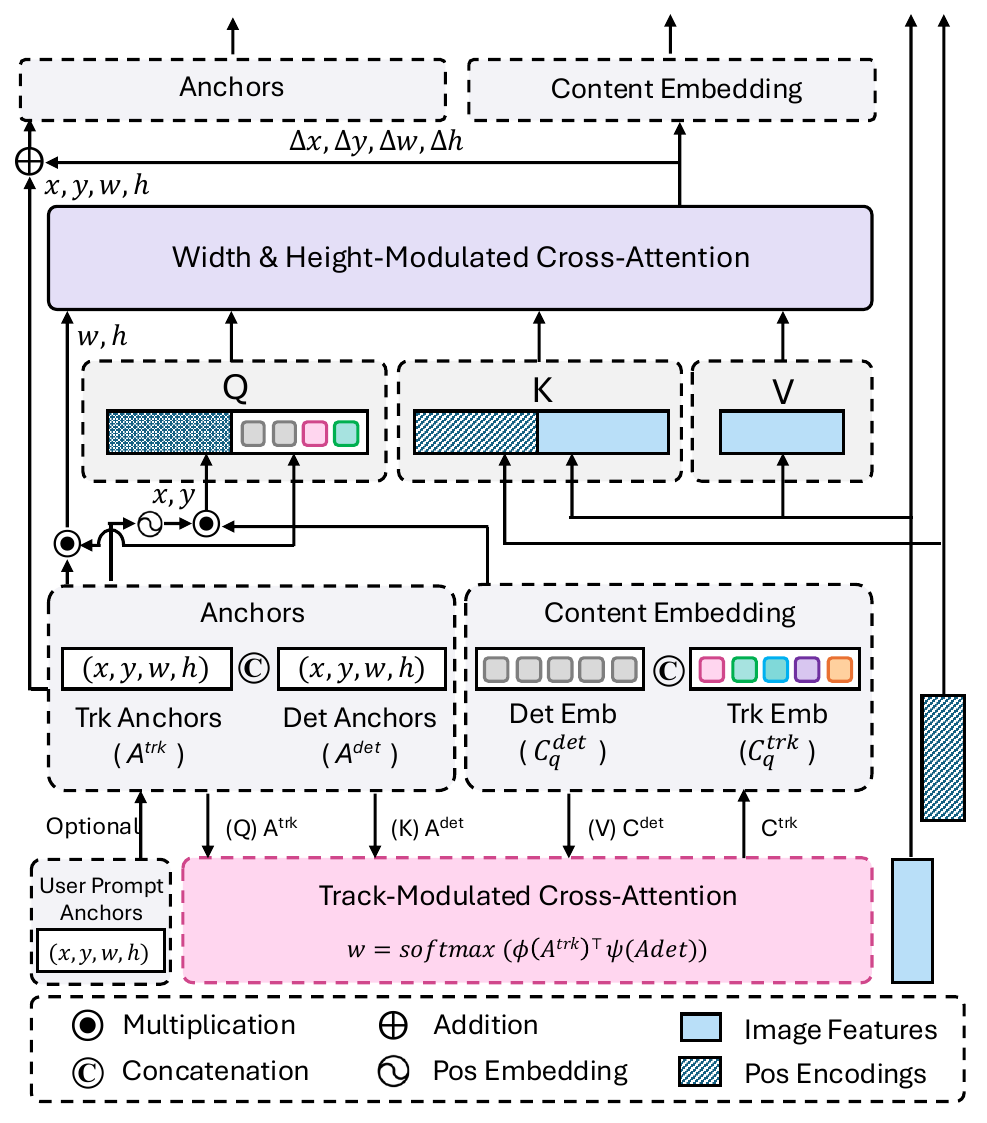}
    \caption{
        Our cross-attention design incorporates detection, tracking, and prompt queries.
        Each query consists of a content embedding and an anchor (learned or provided) defining its positional embedding.
        TMCA first refreshes tracking/prompt queries from current detections, followed by width- and height-modulated cross-attention over image features.
    }
    \label{fig:cross_attention}
    \vskip -2em
\end{wrapfigure}

Given the encoded frame tokens and the sets of detection and tracking queries, we introduce DETRAM, our unified transformer architecture for end-to-end detection, tracking, and human mesh recovery.  
Its core contribution is a cross-attention mechanism that jointly reasons over detection, tracking, and prompt queries within a single decoder, enabling identity-consistent multi-person reconstruction, without requiring any external detector or a separate learned tracking module.
A schematic of our design is provided in \autoref{fig:cross_attention}.

\parahead{Shared Decoder for Detection, Tracking, and Prompt Queries}
To allow detection and tracking queries to interact seamlessly during decoding, we first concatenate their anchor boxes and content embeddings to form query representations.  
Detection queries use learned anchors and content embeddings, while tracking queries use propagated anchors with TMCA-refreshed contents and prompt queries are initialized from user-provided anchors.
Importantly, this concatenation does not merge them into a single query representation. Rather, they are decoded in parallel within the same decoder.
Positional encodings for the combined queries are then generated by applying an MLP to sinusoidal embeddings of the anchor parameters:
\vspace{-3pt}
\[
\begin{aligned}
A_q &= \mathrm{Cat}(A^{\mathrm{det}}_q,\, A^{\mathrm{trk}}_q), \qquad
C_q = \mathrm{Cat}(C^{\mathrm{det}}_q,\, C^{\mathrm{trk}}_q), \\[4pt]
P^{(\cdot)}_{q} &= \mathrm{MLP}\!\big(\mathrm{PE}(A^{(\cdot)}_{q})\big), \\[4pt]
\mathrm{PE}(A) &= 
\mathrm{Cat}\!\big(\mathrm{PE}(x),\, \mathrm{PE}(y),\, 
\mathrm{PE}(w),\, \mathrm{PE}(h)\big),
\end{aligned}
\]
where $\mathrm{PE}(\cdot)$ maps a scalar to a $\tfrac{D}{2}$-dimensional sinusoidal embedding.  
This formulation allows detection, tracking, and prompt queries to interact within a shared decoder while preserving their distinct supervision. Joint decoding enables tracking queries to attend to detection queries, allowing identity information to flow between existing tracks and newly detected instances within the decoder.

\parahead{Decoder Architecture}
Each transformer decoder layer contains a self-attention module followed by a cross-attention module.  
While both require queries, keys, and values, they differ in how these components are constructed.

In the \textit{self-attention} module, all queries, keys, and values use the same content embeddings, while positional encodings are added to the queries and keys:
\vspace{-3pt}
\begin{equation*}
Q_q = C_q + P_q, \qquad 
K_q = C_q + P_q, \qquad 
V_q = C_q .
\label{eq:self_attn}
\end{equation*}

\parahead{Conditional Cross-Attention}
For cross-attention, we adopt a Conditional-DETR–style formulation~\cite{meng2021conditional} that blends semantic and spatial cues to enhance spatial selectivity.  
This yields spatially conditioned queries and keys:
\vspace{-3pt}
\begin{equation*}
\begin{aligned}
Q_q &= \mathrm{Cat}\big(C_q,\, \mathrm{PE}(x_q, y_q)\cdot\mathrm{MLP}^{(\mathrm{csq})}(C_q)\big), \\
K_{x,y} &= \mathrm{Cat}\big(F_{x,y},\, \mathrm{PE}(x,y)\big), \\
V_{x,y} &= F_{x,y},
\end{aligned}
\label{eq:cross_attn}
\end{equation*}
where $F_{x,y} \in \mathbb{R}^D$ denotes the encoder feature at spatial location $(x,y)$, and $\mathrm{MLP}^{(\mathrm{csq})}$ modulates positional embeddings based on semantic content.  
This modulation strengthens the model’s ability to correlate query hypotheses with local image evidence, thereby improving both detection precision and identity stability.

\subsection{Adding New Humans \& Memory Mechanism}

We maintain persistent tracking queries with a simple, non-learned track management heuristic on decoder outputs (i.e., not a separate learned tracker) to initialize and update tracks over time.
DETRAM also optionally uses a lightweight, parameter-free memory bank as temporal regularization to stabilize identity features over long sequences.

\parahead{Initializing and Updating Tracks}
At the first frame, all detections with confidence above $\tau_{\text{det}}$ are instantiated as tracking queries and stored in memory.  
For subsequent frames, a detection is promoted to a new tracking query if its confidence exceeds $\tau_{\text{det}}$ and it has low overlap with all existing tracks (IoU $< \tau_{\text{iou}}$).  
Each tracking query remains active as long as its confidence stays above $\tau_{\text{det}}$; if it becomes inactive for more than $L_{\text{tol}}$ frames, it is removed from memory.
To further manage the life-cycle of tracks, we mark individual tracks as active/inactive/removed based on the following process:
(i) tracks with confidence below $\tau_{\text{det}}$ are marked inactive, and updates to their queries and memory entries are frozen until re-association,
(ii) inactive tracks are resumed upon confidence (or association-score) recovery, and
(iii) inactive tracks are removed if they remain inactive for more than $L_{\text{tol}}$ frames.
This mechanism allows the system to dynamically grow or prune the active set of identities, enabling robust handling of entering, exiting, or reappearing individuals.

\parahead{Key–Value Memory for Long-Term Propagation}
To ensure that tracking queries retain identity cues even under occlusion or pose changes, we adopt a simple yet effective key–value memory mechanism inspired by XMem~\cite{cheng2022xmem} and STCN~\cite{cheng2021rethinking}.  
Let the memory store keys $m_k \in \mathbb{R}^{B \times (N_t \times M) \times D}$ from previous frames and current query keys $q_k \in \mathbb{R}^{B \times N_t \times D}$.  
We measure their pairwise similarity by
$
S = \frac{2\, m_k^\top q_k \;-\; \|m_k\|^2}{\sqrt{C_K}},
$
where $B$ is the batch size, $N_t$ is the number of active tracks, $M$ is the memory length (number of stored slots), and $D$ is the feature dimension.  
Each similarity map $S \in \mathbb{R}^{B \times N_t \times M}$ captures how strongly each tracking query attends to previously stored frames.
We compute attention weights via a top-$k$ softmax,
$
\alpha_{b,n,m} = \mathrm{softmax}_m\!\left(S'_{b,n,m}\right),
$
which focuses retrieval on the most relevant memory slots.  
The final aggregated memory feature is obtained as
$
\mathrm{mem}_{b,n} = \sum_m \alpha_{b,n,m} \, m_v(b,m),
$
where $m_v$ denotes the corresponding memory values.  
This differentiable retrieval mechanism introduces no additional parameters yet provides temporally stable identity features, enabling DETRAM to maintain consistent tracks through occlusions, rapid motion, and long sequences.
The memory bank stores $M$ feature slots per active track, scaling as $O(N_tMD)$ rather than with video length due to the track lifecycle policy.

\subsection{Losses}
Our training objective integrates detection, reconstruction, and tracking supervision within a unified formulation.  
Following DETR and DN-DETR~\cite{carion2020end, li2022dn}, we use Hungarian Matching to associate predictions with ground truth (GT) instances using a combination of bounding boxes, projected joints, and confidence scores.  
We additionally incorporate a denoising mechanism to stabilize learning in the early stages.  
The total loss is expressed as a weighted sum of three components; the matching loss $\mathcal{L}_{\text{matching}}$, the denoising loss $\mathcal{L}_{\text{dn}}$, and the tracking loss $\mathcal{L}_{\text{trk}}$:
\vspace{-3pt}
\begin{equation*}
\mathcal{L} =
\lambda_{\text{matching}}\mathcal{L}_{\text{matching}} +
\lambda_{\text{dn}}\mathcal{L}_{\text{dn}} +
\lambda_{\text{trk}}\mathcal{L}_{\text{trk}}.
\end{equation*}

Further details regarding the loss terms are provided in the supplementary.

\section{Experimental Results}
\parahead{Datasets}
Our model is trained on several multi-person datasets, including AGORA~\cite{patel2021agora}, BEDLAM~\cite{Black_CVPR_2023}, COCO~\cite{lin2014microsoft}, and 3DPW~\cite{vonMarcard2018}, and PoseTrack~\cite{doering2022posetrack21,iqbal2017cvpr,andriluka2018posetrack}.  Among these datasets, AGORA, BEDLAM, and 3DPW provide high-quality 3D mesh annotations that are directly used for supervision. 
For COCO, we rely on pseudo 3D annotations generated by NeuralAnnot~\cite{moon2022neuralannot} and supervise only the projected 2D joints. 
For PoseTrack, we relabel the data with 3D annotations using the Neural Localizer Fields (NLF) method~\cite{sarandi2024nlf}.

\parahead{Evaluation Protocol}
To ensure a fair comparison with existing approaches, we evaluate our framework on PoseTrack~\cite{iqbal2017cvpr,andriluka2018posetrack}, MuPoTS-3D~\cite{mehta2018multi}, BEDLAM~\cite{Black_CVPR_2023}, and 3DPW~\cite{vonMarcard2018}.  Following standard practices~\cite{zheng20213d,kocabas2021pare,pavlakos2019expressive}, we compute both Mean Vertex Error (MVE) and Mean Per-Joint Position Error (MPJPE), reported before and after applying Procrustes Alignment (PA). All errors, unless otherwise specified, are reported in millimeters (mm). 

For the tracking evaluation, we follow the official PoseTrack21~\cite{doering2022posetrack21} protocol and report standard multi-object tracking metrics, including Multi-Object Tracking Accuracy (MOTA), 
which combines false positives (FP), missed detections (FN), and ID switches (IDs) into a single accuracy measure.  We also compute IDF1, ID Precision (IDP), and ID Recall (IDR) to assess the consistency of identity assignment across frames. We also report HOTA (Higher Order Tracking Accuracy)~\cite{luiten2021hota}, which jointly measures detection and association accuracy.  Higher values indicate more stable temporal associations between predicted tracks and ground-truth trajectories.

\parahead{Computational Cost and Implementation details}
We train our model using eight NVIDIA A100 GPUs with a batch size of 2 per GPU. 
For evaluating computational efficiency, we measure the average inference time (in seconds) on a single NVIDIA RTX~A4000 GPU.
We use a perspective camera model, with a fixed fallback camera setting (fixed FOV and centered principal point).
We use a two-stage training protocol: first image-based training on AGORA, BEDLAM, COCO, PoseTrack, and 3DPW, then video training with multi-frame clips (e.g. $T=4$ or $8$) on datasets with sequential annotations (e.g. BEDLAM, PoseTrack, and 3DPW).
At inference time, the memory bank scales with the number of active tracks, $O(N_tMD)$, rather than video length, requiring only about 12--24 KB per active track (in FP32 format).
For video training, we use only temporally contiguous GT track segments and exclude clips with missing/non-continuous GT identities.
Additional implementation details can be found in the supplementary.

\subsection{Evaluation on PoseTrack21 and MuPoTS-3D}

\parahead{Tracking performance} We evaluate our method on the PoseTrack21~\cite{doering2022posetrack21} and MuPoTS-3D~\cite{mehta2018multi} benchmark to assess multi-person tracking performance. 
As shown in \autoref{tab:pose_track_results}, our method achieves the highest overall performance on PoseTrack21, obtaining a MOTA of 73.6 and an IDF1 of 80.6,  while also reducing false negatives (FN) compared to previous approaches such as 4DHumans~\cite{goel2023humans} and CoMotion~\cite{newell2025comotion}.
We also achieve state-of-the-art results on MuPoTS-3D~\cite{mehta2018multi} with a MOTA of 95.9, IDF1 of 97.4 and HOTA of 71.3.
These improvements indicate better identity preservation and temporal associations. 

\begin{figure}[t]  
    \centering
    \vskip -2mm
    \includegraphics[width=1.0\linewidth]{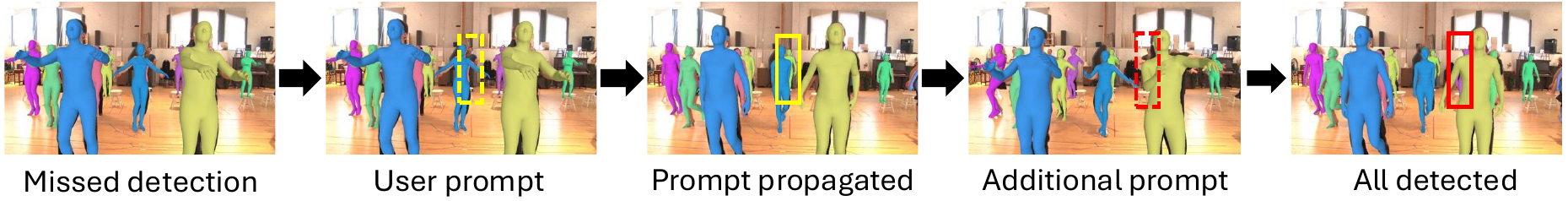}
    \vskip -2mm
    \caption{
    \textbf{\ours + prompts.}
    User prompts (yellow/red dashed boxes) can be specified when false negatives (missed detections) occur. Propagated boxes (solid line boxes) then follow the person in future frames, recovering all humans.
    }
    \label{fig:prompting}
    \vskip -2em
\end{figure}

As \ours allows prompting, we simulate a realistic semi-automatic setting to quantify the potential of this capability: whenever a missed detection (false negative) first occurs, we pass in a bounding box derived from the ground-truth annotation for that person (an example is demonstrated in \autoref{fig:prompting}).
While this oracle protocol provides an upper bound on the benefit of prompting, the consistent improvements across all tracking metrics (last row of \autoref{tab:pose_track_results} and \autoref{tab:mupots}) validate that prompt queries are effectively propagated by the model.
Beyond tracking accuracy, \ours runs at 10.87 FPS on PoseTrack21 -- 1.9× faster than CoMotion~\cite{newell2025comotion} and over 21× faster than 4DHumans~\cite{goel2023humans}, highlighting the efficiency of our single-stage architecture.

\begin{table*}[t]
\centering
\caption{Evaluation of tracking performance on the PoseTrack21 (left) and MuPoTS-3D (right) benchmarks.
\emph{``\ours + prompts''} simulates a user-prompting setting where false negatives are specified via box prompts based on ground-truth, leading to improvements across all tracking metrics.
%
}
\setlength{\tabcolsep}{4pt}
\begin{subtable}{0.57\textwidth}
    \vskip -6mm
    \caption{PoseTrack21}
    \vskip -3mm
    \renewcommand{\arraystretch}{1.2}
    \resizebox{1.0\textwidth}{!}{%
    \begin{tabular}{lccccccc}
        \toprule
        Method & MOTA$\uparrow$ & IDF1$\uparrow$ & IDP$\uparrow$ & IDR$\uparrow$ &
        FP$\downarrow$ & FN$\downarrow$ & FPS$\uparrow$ \\
        \midrule
        {\footnotesize With original evaluation method} \\
        TRMOT~\cite{wang2020towards} & 47.2 & 57.3 & 70.0 & 46.6 & -- & -- & -- \\
        FairMOT~\cite{zhang2021fairmot} & 56.3 & 63.2 & 81.0 & 51.8 & -- & -- & -- \\
        CorrTrack + ReID~\cite{doering2022posetrack21} & 52.0 & 66.5 & 72.4 & 61.4 & -- & -- & -- \\
        Tracktor++~\cite{bergmann2019tracking} & 59.5 & 69.3 & 76.4 & 63.5 & -- & -- & -- \\
        CoMotion ~\cite{newell2025comotion} & \underline{67.6} & \underline{77.9} & \textbf{83.4} & \underline{73.0} & -- & -- & -- \\
        DETRAM (Ours) & \textbf{71.0} & \textbf{78.3} & \underline{81.6} & \textbf{74.3} & -- & -- & -- \\
        \midrule
        {\footnotesize With ignore region fix from~\cite{newell2025comotion}} \\
        4DHumans~\cite{goel2023humans} (reproduced) & 56.7 & 70.9 & \textbf{87.1} & 59.7 & \textbf{7817} & 50652 & 0.51\\
        CoMotion~\cite{newell2025comotion} & \underline{71.4} & \underline{79.5} & \textbf{87.1} & \underline{73.0} & 8115 & \underline{30394} & \underline{5.68} \\
        DETRAM (Ours)  & \textbf{73.6} & \textbf{80.6} & \underline{85.0} & \textbf{75.1} & \underline{8066} & \textbf{26620} & \textbf{10.87}\\
        \midrule
        DETRAM (Ours) $+$ prompts & \textbf{75.7} & \textbf{81.8} & \underline{85.0} & \textbf{77.2} & \underline{8059} & \textbf{23369} & \textbf{10.87} \\
        \bottomrule
    \end{tabular}
    }
    \label{tab:pose_track_results}
\end{subtable}
\hfill
\begin{subtable}{0.42\textwidth}
    \vskip -6mm
    \caption{MuPoTS-3D}
    \vskip -3mm
    \renewcommand{\arraystretch}{1.44}
    \resizebox{1.0\textwidth}{!}{%
    \begin{tabular}{lcccc}
        \toprule
        Method & IDs$\downarrow$ & MOTA$\uparrow$ & IDF1$\uparrow$ & HOTA$\uparrow$ \\
        \midrule
        Trackformer~\cite{meinhardt2022trackformer} & 43 & 24.9 & 62.7 & 53.2 \\
        Tracktor~\cite{bergmann2019tracking}        & 53 & 51.5 & 70.9 & 50.3 \\
        FlowPose~\cite{xiu2018poseflow}             & 49 & 21.4 & 67.1 & 41.8 \\
        T3DP~\cite{rajasegaran2021tracking}         & 38 & 62.1 & 79.1 & 59.2 \\
        PHALP~\cite{rajasegaran2022tracking}        & 22 & 66.2 & 81.4 & 59.4 \\
        ByteTrack~\cite{zhang2022bytetrack}         & 15 & 73.3 & 84.6 & 63.5 \\
        TRACE~\cite{sun2023trace}                   & \textbf{0} & 86.9 & 93.4 & 65.3 \\
        CoMotion~\cite{newell2025comotion}          & \textbf{0} & 92.1 & 95.9 & 69.2 \\
        \ours (Ours)                                & \textbf{0} & \textbf{95.9} & \textbf{97.4} & \textbf{71.3} \\
        \midrule
        \ours (Ours) $+$ prompts & \textbf{0} & \textbf{96.5} & \textbf{98.2} & \textbf{71.6} \\
        \bottomrule
    \end{tabular}
    }
    \label{tab:mupots}
\end{subtable}
\end{table*}

\begin{figure*}[t]  
    \centering
    \includegraphics[width=1.0\linewidth]{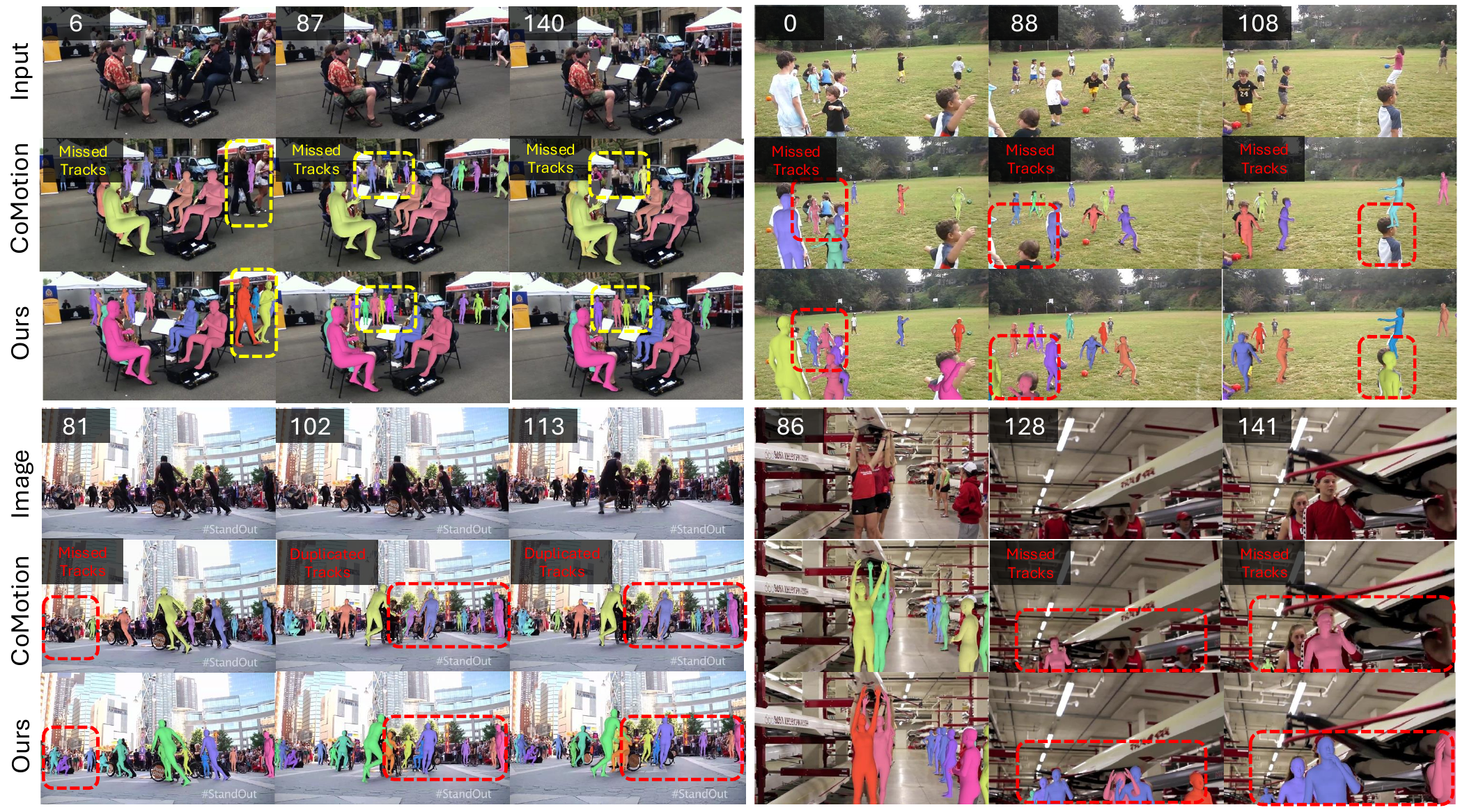}
    \vskip -3mm
    \caption{
       Qualitative results on the PoseTrack21 dataset show that we are able to detect more humans with fewer false negatives and false positives compared to CoMotion. The numbers in the grey boxes denote the frame index, while the description in grey boxes along with dotted boxes highlight the differences between CoMotion and \ours.
    }
    \label{fig:qual_posetrack}
    
\end{figure*}

\parahead{Qualitative results} \autoref{fig:qual_posetrack} presents qualitative comparisons between our method and CoMotion~\cite{newell2025comotion} on diverse multi-person scenes.  In the first sequence, our model detects more individuals accurately, especially in the background (yellow box). 
In the second and fourth sequences, our approach maintains stable tracking and consistent reconstruction despite camera movement, frequent occlusions, and limited full-body visibility (red boxes). 
The third example highlights that our method produces fewer false positives and maintains better identity consistency across frames. 
Overall, \ours demonstrates more reliable multi-person tracking and reconstruction under challenging real-world conditions.

\vspace{-1em}
\subsection{Evaluation on BEDLAM}
\vspace{-0.5em}

\parahead{Tracking performance} We evaluate our approach on the BEDLAM~\cite{Black_CVPR_2023} benchmark following the evaluation protocol established by MaQ~\cite{liu2025motions}. 
As shown in \autoref{tab:bedlam_comparison}, our method achieves the best tracking accuracy, achieving a MOTA of 99.18 and an IDF1 of 99.37, while drastically reducing the number of ID switches to only 37.  These results highlight the model’s capability to maintain identity consistency and temporal coherence across long and crowded sequences, outperforming all baselines by a clear margin.

\begin{table*}[t]
\centering
\caption{Evaluation of tracking and reconstruction performance on the BEDLAM dataset.
We significantly out-perform the previous methods in key tracking metrics (e.g. IDs and IDF1) and reconstruction metrics (e.g. MPJPE) while attaining comparable inference time and HOTA to MaQ~\cite{liu2025motions} and comparable PA-MPJPE to CoMotion~\cite{newell2025comotion}. 
\label{tab:bedlam_comparison}
}
\vskip -6mm
\resizebox{\textwidth}{!}{
\setlength{\tabcolsep}{4pt}
\renewcommand{\arraystretch}{1.2}
\begin{tabular}{l|cccccccC{17mm}}
\toprule
Method & IDs$\downarrow$ & MOTA$\uparrow$ & IDF1$\uparrow$ & HOTA$\uparrow$ & 
MPJPE$\downarrow$ & PA-MPJPE$\downarrow$ & PVE$\downarrow$ & Inference Time$\downarrow$ \\
\midrule
Fast-RCNN \cite{girshick2015fast} + SMPLer-X \cite{cai2023smpler} + ByteTrack \cite{zhang2022bytetrack} 
& 1660 & 78.83 & 81.09 & 70.41 & 119.19 & 52.62 & 129.22 & 0.463 \\
Multi-HMR \cite{baradel2024multi} + ByteTrack \cite{zhang2022bytetrack} 
& 515 & 91.35 & 92.54 & 84.59 & 81.58 & 42.93 & 88.88 & 0.201 \\
MaQ~\cite{liu2025motions}
& \underline{128} & 95.15 & 97.01 &\textbf{97.73} & 79.88 & 45.56 & 86.40 & \textbf{0.027} \\
CoMotion~\cite{newell2025comotion}
& 885 & \underline{95.22} & \underline{97.39} & 70.94 & \underline{55.40} & \textbf{24.50} & \underline{62.17} & 0.239 \\
\midrule
\ours (Ours) & \textbf{37} & \textbf{99.18} & \textbf{99.37} & \underline{95.90} & \textbf{48.68} & \underline{31.24} & \textbf{56.33} & \underline{0.069} \\




\bottomrule
\end{tabular}
}
\end{table*}

\parahead{Reconstruction performance} In addition to tracking, our method significantly improves 3D reconstruction quality.  It achieves lower MPJPE of 48.68~mm, PA-MPJPE of 31.24~mm,  and PVE of 56.33~mm compared to previous approaches.  This demonstrates the effectiveness of our unified architecture,  which jointly optimizes detection, tracking, and 3D reconstruction in a single framework.  
Although our method achieves superior tracking and reconstruction accuracy overall, the HOTA score is slightly lower than MaQ~\cite{liu2025motions}.
We believe this to be due to the differences how 2D bounding boxes are defined, which can reduce HOTA even though the underlying 3D pose and shape estimations are more accurate and consistent across frames.
Furthermore, while we out-perform CoMotion~\cite{newell2025comotion} on most metrics, it shows a low PA-MPJPE score, which may be due to its larger training datasets (which include DanceTrack and InstaVariety, which we do not train on).

\begin{figure*}[h]  
    \centering
    \vskip -3mm
    \includegraphics[width=1.0\linewidth]{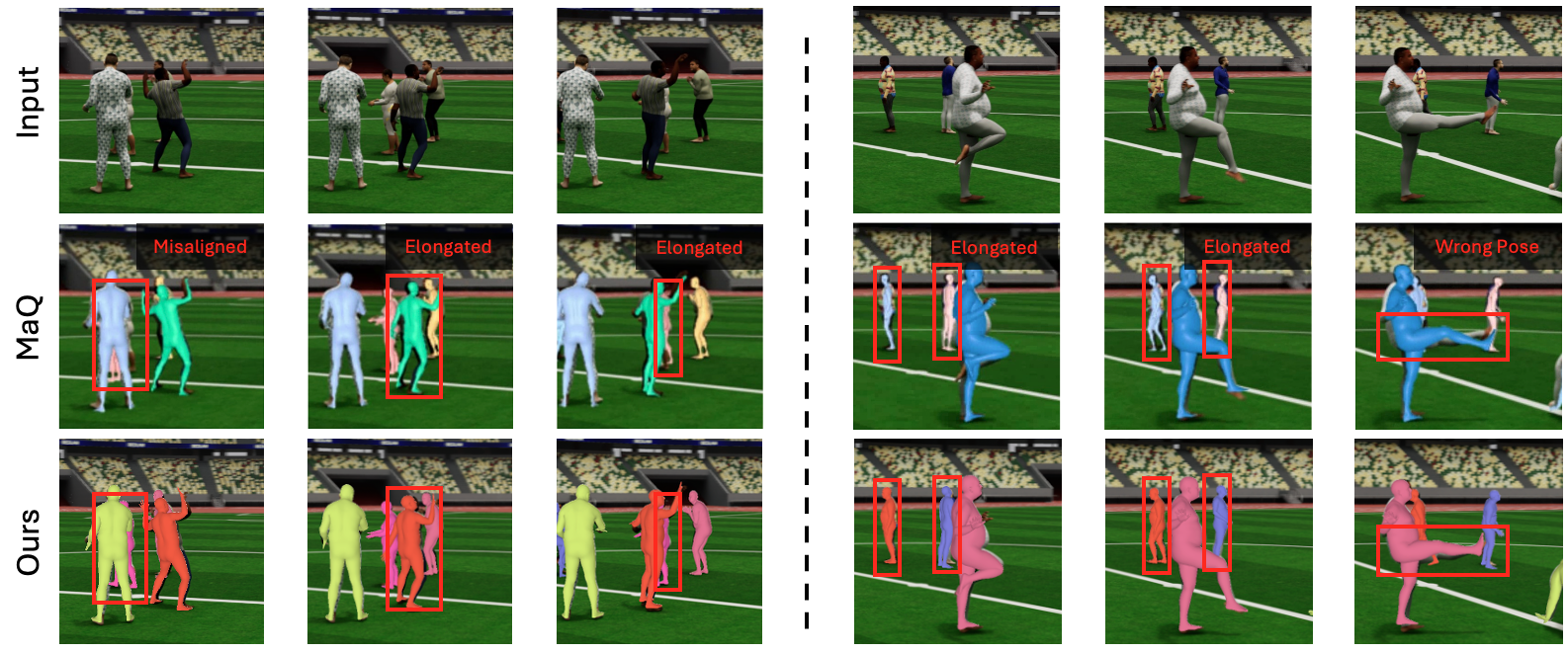}
    \vskip -3mm
    \caption{
       Qualitative results on BEDLAM, in comparison with MaQ~\cite{liu2025motions}. Note that the MaQ visualizations are taken directly from their paper as they do not provide inference code. 
       \ours predicts more accurate human meshes (pose and proportions). 
    }
    \label{fig:qual_bedlam}
\end{figure*}

\parahead{Qualitative results}
\autoref{fig:qual_bedlam} shows qualitative comparisons with MaQ~\cite{liu2025motions} on the BEDLAM dataset.  Our method demonstrates robust performance even under occlusions, successfully reconstructing multiple interacting people with consistent identities.  Compared to MaQ, our approach generates more anatomically plausible human shapes and poses, particularly preserving body proportions.  In contrast, MaQ tends to produce overly slim or elongated body shapes.  These qualitative observations further validate the advantage of our unified architecture in achieving both accurate 3D pose estimation and stable multi-person tracking.

\vspace{-0.2em}
\subsection{Evaluation on 3DPW}
\vspace{-0.2em}
\parahead{Tracking and reconstruction performance}
We further evaluate \ours on the Dyna3DPW and 3DPW dataset by following the evaluation protocol in~\cite{sun2023trace, liu2025motions}. 
As shown in \autoref{tab:threedpw}, \ours achieves strong tracking consistency, obtaining the best MOTA (99.8), IDF1 (99.9), and HOTA (79.1), while maintaining zero ID switches. In terms of reconstruction, our method achieves competitive accuracy with an MPJPE of 60.0~mm and PVE of 70.1~mm, outperforming prior methods in overall tracking stability while maintaining 3D reconstruction quality. 
These results demonstrate that our identity-consistent query formulation generalizes well to in-the-wild videos with camera motion.
CoMotion achieves a lower PA-MPJPE (37.3 vs. 38.3 mm), likely due to additional training data (DanceTrack and InstaVariety). However, \ours outperforms it on MPJPE (60.0 vs. 60.6), PVE (70.1 vs. 71.2), and all tracking metrics.

\begin{table}[t]
\centering
\caption{Evaluation of tracking and reconstruction performance on Dyna3DPW and 3DPW, respectively, following the protocol in \cite{sun2023trace, liu2025motions}.}
\label{tab:threedpw}
\vskip -3mm
\resizebox{1.0\textwidth}{!}{
\setlength{\tabcolsep}{4pt}
\renewcommand{\arraystretch}{1.2}
\begin{tabular}{l|cccc|ccc}
\hline
\multirow{2}{*}{Method} & \multicolumn{4}{c|}{Dyna3DPW (Tracking)} & \multicolumn{3}{c}{3DPW (Reconstruction)} \\
& IDs$\downarrow$ & MOTA$\uparrow$ & IDF1$\uparrow$ & HOTA$\uparrow$
      & PA-MPJPE$\downarrow$ & MPJPE$\downarrow$ & PVE$\downarrow$ \\
\hline
BEV~\cite{sun2022putting} + ByteTrack~\cite{zhang2022bytetrack} & 37 & 93.6 & 79.1 & 59.3 & 46.9 & 78.5 & 92.3 \\
TRACE~\cite{sun2023trace}                 & 1 & 99.3 & 99.7 & 74.7 & 50.8 & 80.3 & 98.1 \\
MaQ~\cite{liu2025motions}                 & \textbf{0} & 99.6 & 99.8 & 70.8 & 44.7 & 72.6 & 84.9 \\
CoMotion~\cite{newell2025comotion}        & \textbf{0} & 95.7 & 97.8 & 72.8 & \textbf{37.3} & 60.6 & 71.2 \\
\ours (Ours)                              & \textbf{0} & \textbf{99.8} & \textbf{99.9} & \textbf{79.1} & 38.3 & \textbf{60.0} & \textbf{70.1} \\
\hline
\end{tabular}
}
\end{table}


\begin{table}[t]
\begin{minipage}[t]{0.59\linewidth}
    \centering
    \caption{Ablation study comparing the effects of the tracking query (Track Q) and memory mechanism (Mem Bank) on PoseTrack21 and BEDLAM.
    }
    \vskip -3mm
    \resizebox{\textwidth}{!}{
        \setlength{\tabcolsep}{4pt}
        \renewcommand{\arraystretch}{1.2}
        \begin{tabular}{c c | c c | c c c c}
        \toprule
        \multirow{2}{*}{Track Q} & \multirow{2}{*}{Mem Bank} &
        \multicolumn{2}{c|}{PoseTrack21} & 
        \multicolumn{4}{c}{BEDLAM} \\
         &  & MOTA$\uparrow$ & IDF1$\uparrow$ &
         MOTA$\uparrow$ & IDF1$\uparrow$ & IDs$\downarrow$ & MPJPE$\downarrow$ \\
        \midrule
        \checkmark & \checkmark & \textbf{73.6} & \textbf{80.6} & \textbf{99.18} & \textbf{99.37} & \underline{37} & 48.68 \\
        \checkmark & -          & \underline{73.3} & \underline{80.3} & \underline{99.10} & \underline{99.33} & \textbf{32} & \underline{47.33} \\
        - & \checkmark                  & 68.9 & 78.1 & 98.19 & 98.90 & \underline{37} & 48.27 \\
        - & -                  & 68.3 & 77.1 & 96.70 & 97.98 & 58 & \textbf{46.91} \\
        \bottomrule
        \end{tabular}
        \label{tab:ablation_trackq_mem_video}
    }
\end{minipage}
\hfill
\begin{minipage}[t]{0.39\linewidth}
    \centering
    \caption{Stratified analysis of tracking performance on crowded scenes from PoseTrack21.}
    \vskip -3mm
    \resizebox{\textwidth}{!}{
        \setlength{\tabcolsep}{4pt}
        \renewcommand{\arraystretch}{1.8}
        \begin{tabular}{c|c|cc|cc}
        \hline
        \multirow{2}{*}{GT Bin} & \multirow{2}{*}{\#Seq}
        & \multicolumn{2}{c|}{\ours (Ours)}
        & \multicolumn{2}{c}{CoMotion~\cite{newell2025comotion}} \\
         & 
        & IDF1 $\uparrow$ & MOTA $\uparrow$
        & IDF1 $\uparrow$ & MOTA $\uparrow$ \\
        \hline
        1--10  & 103 & 87.69 & \textbf{82.13} & \textbf{87.72} & 82.05 \\
        11--20 & 44  & \textbf{79.58} & \textbf{74.60} & 75.81 & 69.72 \\
        21+    & 23  & \textbf{71.66} & \textbf{66.83} & 70.67 & 60.53 \\
        \hline
        \end{tabular}
        \label{tab:crowded}
    }
\end{minipage}
\vskip -3mm
\end{table}

\vspace{-0.2em}
\subsection{Further Analysis}
\vspace{-0.2em}
\parahead{Ablation Study}
In \autoref{tab:ablation_trackq_mem_video} we conduct an ablation study. We show that applying \emph{tracking queries} (Track Q) and the \emph{memory mechanism} (Mem Bank) independently results in modest improvements in tracking performance (MOTA, IDF1), while slightly sacrificing reconstruction performance (MPJPE). 
Accordingly, we treat the memory mechanism as an optional, parameter-free temporal regularization that stabilizes identity features over long sequences.
The memory bank yields a modest gain in MOTA on PoseTrack21 (73.3 to 73.6) with a small runtime cost (11.21 to 10.87 FPS). Its higher BEDLAM MPJPE (47.33 to 48.68) but improved MuPoTS-3D 3DPCK@150~\cite{mehta2018single} (95.66\% to 97.84\%) suggests a recall--coverage trade-off. That is, recovering more occluded people can increase the number of difficult-to-reconstruct instances.
While its individual effect is modest, combining it with tracking queries yields the best overall tracking performance on both PoseTrack21 and BEDLAM. This is consistent with our design intent: the memory bank is not a primary contribution but a complementary component that provides diminishing returns when tracking queries already maintain strong identity
signals.
The performance differences are intuitive, as \emph{tracking queries} aid in maintaining identity continuity and can track more individuals than detection queries in crowded scenes, while the \emph{memory mechanism} continues tracking known entities. This is why \emph{tracking queries} are particularly impactful on the PoseTrack21 dataset, which contains numerous entities and entry/exit events. Tracking queries alone produce slightly fewer ID switches on BEDLAM (32 vs. 37), likely due to occasional memory-induced drift from stale embeddings after long occlusions. However, the combined configuration achieves higher overall MOTA and IDF1, indicating that the memory bank’s benefits outweigh occasional misassociations.

\parahead{Crowded Scenes}
Challenges in multi-person tracking often arise in crowded scenes.
In \autoref{tab:crowded}, we perform a comparison of \ours against CoMotion~\cite{newell2025comotion} by evaluating tracking performance stratified by the number of people present in a given sequence.
We find that in sequences with few people (1--10), \ours and CoMotion perform comparably, while \ours performs notably better with increasing crowdedness (11--20 and 21+ bins), exhibiting over 10\% improvement in MOTA over CoMotion for the most crowded sequences (21+).

\begin{table}[t]
\centering
\caption{
Ablation of TMCA and content-conditioned spatial modulation.
}
\label{tab:tmca_ablation}
\scriptsize
\setlength{\tabcolsep}{3.5pt}
\begin{tabular}{lcc|cccc}
\toprule
\multirow{2}{*}{Variant} 
& \multicolumn{2}{c|}{PoseTrack21} 
& \multicolumn{4}{c}{BEDLAM} \\
\cmidrule(lr){2-3} \cmidrule(lr){4-7}
& MOTA$\uparrow$ & IDF1$\uparrow$ 
& MOTA$\uparrow$ & IDF1$\uparrow$ & IDs$\downarrow$ & MPJPE$\downarrow$ \\
\midrule
w/o Track-Q        & 68.3 & 77.1 & 96.70 & 97.98 & 58  & 46.91 \\
TMCA $\rightarrow$ direct & 71.7 & 79.6 & 99.16 & 99.35 & 47  & 49.37 \\
$\mathrm{MLP}^{(\mathrm{csq})} \rightarrow$ id. & 70.2 & 77.1 & 98.38 & 98.05 & 305 & 52.73 \\
\ours~(full)       & 73.6 & 80.6 & 99.18 & 99.37 & 37  & 48.68 \\
\bottomrule
\end{tabular}
\vspace{-0.5em}
\end{table}

\parahead{TMCA and query modulation ablations}
As shown in \autoref{tab:tmca_ablation}, replacing TMCA with direct propagation reduces PoseTrack21 MOTA from 73.6 to 71.7, confirming the benefit of refreshing tracking queries from current detections. Replacing $\mathrm{MLP}^{(\mathrm{csq})}$ with an identity mapping increases BEDLAM ID switches from 37 to 305 and worsens MPJPE from 48.68 to 52.73, showing that content-conditioned spatial modulation is critical for stable identity tracking. Overall, TMCA-based tracking trades slightly higher MPJPE for stronger identity consistency.


\vspace{-0.5em}
\section{Conclusion}
\vspace{-0.7em}
We introduced \ours, a unified transformer framework that jointly performs detection, tracking, and human mesh recovery within a single feed-forward pass per frame. By integrating detection, tracking, and prompt queries, it maintains temporal consistency, handles occlusions, and enables controllable tracking. 
Its framewise design enables online streaming inference with prompt re-initialization.
Experiments on PoseTrack21, MuPoTS-3D, 3DPW, and BEDLAM demonstrate improved accuracy and efficiency. Future work will explore explicit temporal smoothness metrics, motion priors for improved temporal coherence, and additional prompting modalities such as text or point-click inputs.

\vspace{1.0em}
\noindent\textbf{Acknowledgments.}
This work was supported by NSF grant CRCNS-2309041 and the Harvard Data Science Initiative Trust in Science Fund Award.

\clearpage  


%
%
\bibliographystyle{splncs04}
\bibliography{main}

\clearpage
\setcounter{page}{1}
\appendix
\counterwithin{figure}{section}
\counterwithin{table}{section}

\parahead{Overview}
This supplementary materials document contains additional information regarding method, dataset, and implementation details as well as additional experimental results of \ours. References to the corresponding sections in the main paper are marked \textcolor{blue}{in blue}.

\section{Method Details}

\subsection{Model Architecture (\textcolor{blue}{Section 3})}


\parahead{Camera Model}
All predictions are made in the camera coordinate system using a pinhole (perspective) projection model with intrinsic matrix $K$. To leverage 2D annotations for supervision, we project 3D joints onto the image plane using focal length $f$ and principal point $(p_u, p_v)$. Given a 3D point $(x,y,z)$, the projected image coordinates $(u,v)$ are
\begin{equation}
u=\frac{f \times x}{z}+p_u, \qquad v=\frac{f \times y}{z}+p_v.
\label{eq:pinhole_projection}
\end{equation}

When dataset-specific camera intrinsics are unavailable, we initialize $K$ using a canonical calibration corresponding to a fixed field of view (FOV) of $60^\circ$~\cite{baradel2024multi}. Let $S_{hr}$ denote the longer side of the image. We define the focal length as
\begin{equation}
f = \frac{S_{hr}}{2\tan(\mathrm{FOV}/2)},
\end{equation}
and set the principal point to the image center, i.e., $(p_u,p_v)=(W/2,H/2)$. 

\parahead{\ours Output Heads and Iterative Refinement}
After self- and cross-attention, each \ours (decoder layer) outputs refined query embeddings, which are fed into lightweight MLP heads to regress root depth, SMPL pose parameters, shape coefficients, 3D and 2D joints, bounding boxes, and detection confidence.  
Following prior work~\cite{hmrKanazawa18, kocabas2021pare, li2022cliff}, predicted SMPL parameters are offset from the mean SMPL pose and shape for stability.  
These predictions are then reintroduced into the next decoding layer, enabling iterative refinement of human localization, identity association, and 3D reconstruction in a unified feed-forward process.

\subsection{Losses (\textcolor{blue}{Section 3.4})}

\parahead{Loss Components}
All three objectives (\emph{matching}, \emph{dn}, and \emph{trk}) share a common loss formulation. They differ only in the GT–prediction pairs from which they are computed.  
Each objective supervises a comprehensive set of outputs using the following decomposition:
\vspace{-3pt}
\begin{multline*}
\mathcal{L}_{(\cdot)} =
\lambda_{\text{depth}}\mathcal{L}_{\text{depth}} +
\lambda_{\text{pose}}\mathcal{L}_{\text{pose}} +
\lambda_{\text{shape}}\mathcal{L}_{\text{shape}} +
\lambda_{\text{j3d}}\mathcal{L}_{\text{j3d}} \\
+ \lambda_{\text{j2d}}\mathcal{L}_{\text{j2d}} +
\lambda_{\text{box}}\mathcal{L}_{\text{box}} +
\lambda_{\text{conf}}\mathcal{L}_{\text{conf}},
\end{multline*}
where $(\cdot) \in \{\text{matching}, \text{dn}, \text{trk}\}$.  
These terms supervise root depth, SMPL pose parameters, shape coefficients, 3D and 2D joints, bounding boxes, and confidence.  
We employ L1 loss or GIoU depending on the target.

\parahead{Matching and Denoising}
The matching loss applies Hungarian assignment to compute supervision for all predicted queries, forming the backbone of our reconstruction and detection training.  
To further enhance robustness, we adopt the DN-DETR denoising strategy~\cite{li2022dn}, injecting noisy perturbations of GT annotations as auxiliary training samples.  
This encourages the model to recover clean predictions from corrupted inputs, significantly accelerating convergence and improving stability.

\parahead{Tracking Supervision}
In contrast to detection queries, where multiple noisy augmentations are used for denoising, we supervise each tracking query using a single GT reference.  
This avoids over-regularizing identity embeddings and ensures that tracking queries learn stable, identity-preserving representations that propagate reliably across frames.

\section{Datasets (\textcolor{blue}{Section 4})}
We train our model on multi-person human pose and mesh datasets. Below, we summarize all datasets used for training and evaluation.

\parahead{3DPW}
3DPW is an outdoor, multi-person (1 or 2 persons), in-the-wild dataset containing video sequences captured. It provides accurate 3D SMPL annotations and includes approximately 17K images for training, 8K for validation, and 24K for testing. Owing to its challenging real-world conditions (e.g., including occlusions, fast motions, and social interactions), it has become a standard benchmark for evaluating 3D human mesh reconstruction models.

\parahead{AGORA}
AGORA is a high-realism, synthetic multi-person dataset consisting of diverse scenes with accurate SMPL and SMPL-X ground truth. It contains roughly 14K images for training, 2K for validation, and 3K for testing, generated from 4,240 high-quality human scans with a wide variety of poses, shapes, and lighting conditions.

\parahead{BEDLAM}
BEDLAM is a large-scale synthetic dataset designed for multi-person human reconstruction. It contains approximately 286K images with 951K person instances for training and 29K images with 96K person instances for validation. The dataset includes significant diversity in body shapes, motions, clothing, hair, and accessories, with each scene containing 1–10 people captured from multiple camera angles. Following prior work, we downsample the training set by a factor of 6 for efficient experimentation. Since the official BEDLAM test set is not publicly available, we follow prior work and evaluate on the validation set instead. In particular, we report results on the subset of 250 validation images commonly used in previous studies \cite{liu2025motions} to ensure a fair comparison.

\parahead{COCO}
COCO is a large-scale real-world multi-person 2D human keypoint dataset and is widely used for pose estimation. We use it to improve the generalization ability of our model on natural images. Due to 3D ambiguity and potential label noise, we only supervise projected 2D joints using pseudo annotations generated by NeuralAnnot~\cite{moon2022neuralannot}. Following common practice, we uniformly downsample COCO by a factor of 4, resulting in 16K images with 66K person instances for training.

\parahead{PoseTrack21}
PoseTrack21 is a multi-person video dataset designed for human pose tracking. It contains long-term person trajectories with 2D joint annotations for each frame. We use the training set, consisting of approximately 11K labeled frames and 153K annotated person instances, to strengthen temporal consistency and improve tracking robustness. The dataset includes challenging crowded scenes, heavy occlusions, and fast motion, making it well suited for evaluating multi-frame pose estimation and identity preservation. Additionally, following common practice, we generate SMPL parameters for each person using Neural Localizer Fields (NLF~\cite{sarandi2024nlf}) and use these pseudo-annotations to provide 3D supervisory signals during training.

\section{Implementation Details}


The encoder and decoder modules are initialized from a pretrained SAT-HMR checkpoint~\cite{su2025sat}. We train our model on 8 NVIDIA A100 GPUs.
We use a two-stage training protocol. For the image-based stage, we follow the same configuration as SAT-HMR and train on AGORA, BEDLAM, COCO, PoseTrack, and 3DPW with a per-GPU batch size of 6. 
For the video-based stage, we train on BEDLAM, 3DPW, and PoseTrack video sequences using multi-frame clips, varying the clip length $T$ between 4 and 8, with a per-GPU batch size of 1.
For video training, we use only datasets with sequential annotations and sample temporally contiguous ground-truth track segments.
We exclude clips in which the target ground-truth identity disappears or is not continuously annotated across the selected frames.
To balance dataset scale, we sample BEDLAM, 3DPW, and PoseTrack with relative proportions of 20\%, 100\%, and 100\%, respectively. We use the AdamW optimizer with a cosine learning-rate schedule, a base learning rate of $2\times10^{-5}$, and $1\times10^{-5}$ for the encoder. Prompt-related components are trained with the same base learning rate. Augmentation includes random rotation, scaling, and horizontal flipping.

\section{Extended Results}
\begin{figure*}[t]
\centering

\begin{minipage}[t]{0.48\textwidth}
\vspace{0pt} 
\resizebox{\linewidth}{!}{
\begin{tabular}{lcc}
\toprule
\multirow{2}{*}{Method} &
\multicolumn{2}{c}{\textbf{PoseTrack}} \\
\cmidrule(lr){2-3}
& PCKn@0.05$\uparrow$ & PCKn@0.1$\uparrow$ \\
\midrule
\multicolumn{3}{l}{\textit{oracle crop}} \\
\midrule
PyMAF~\cite{zhang2021pymaf} & 0.77 & 0.92 \\
CLIFF~\cite{li2022cliff} & 0.75 & 0.92 \\
PARE~\cite{kocabas2021pare} & 0.79 & 0.93 \\
PyMAF-X~\cite{zhang2023pymaf} & 0.85 & 0.95 \\
HMR 2.0a~\cite{goel2023humans} & 0.86 & 0.97 \\
HMR 2.0b~\cite{goel2023humans} & \textbf{0.90} & \textbf{0.98} \\
CoMotion~\cite{newell2025comotion} & 0.90 & 0.97 \\
\midrule
\multicolumn{3}{l}{\textit{full image}} \\
\midrule
CoMotion~\cite{newell2025comotion} & \textbf{0.88} & \textbf{0.96} \\
Ours & 0.84 & \textbf{0.96} \\
\bottomrule
\end{tabular}
}
\end{minipage}
\hfill
\begin{minipage}[t]{0.48\textwidth}
\vspace{0pt} 
\centering
\includegraphics[width=\linewidth]{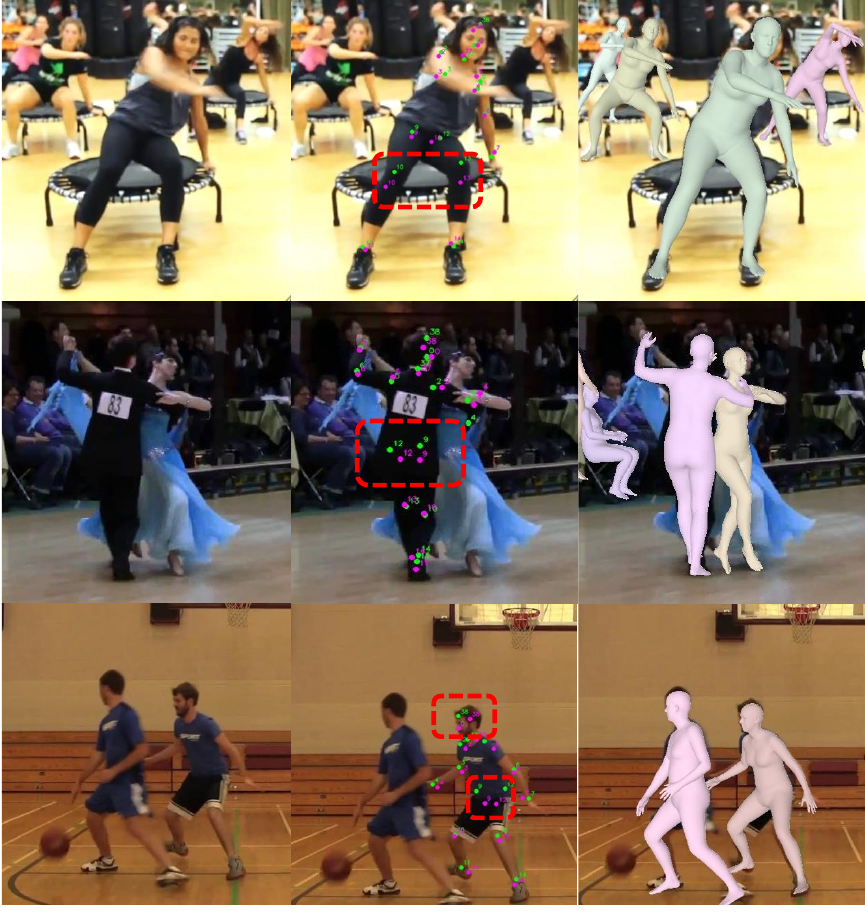}
\end{minipage}

\vskip -2mm
\caption{
\textbf{Left:} Quantitative comparison on PoseTrack using PCKn@0.05 and PCKn@0.1.
\textbf{Right:} Visualization of projected SMPL 2D keypoints (pink) against PoseTrack21 annotations (green). 
Systematic offsets (e.g., pelvis and head) due to structural differences between SMPL joints and PoseTrack21 keypoint definitions can reduce PCK even when meshes appear visually accurate.
}
\label{fig:posetrack_pckn_combined}
\vskip -2em
\end{figure*}

\parahead{Posetrack PCK Evaluation} 
\autoref{fig:posetrack_pckn_combined} reports our evaluation on the PoseTrack21 dataset using the PCKn@0.05 and PCKn@0.1 metrics. Following  CoMotion~\cite{newell2025comotion}, we perform evaluation in the full-image setting without relying on oracle crops. Our method achieves competitive performance. However, the scores are slightly lower than CoMotion’s results. This discrepancy primarily arises from the annotation format mismatch between SMPL joints (predicted by our model) and the 2D PoseTrack21 keypoints, which follow a different skeletal definition and joint localization convention. Our model was trained by SMPL based datasets. As a result, direct comparison introduces inherent structural differences that naturally affect PCK-based evaluation.

Moreover, as illustrated in \autoref{fig:posetrack_pckn_combined}, even when our method produces accurate SMPL meshes, the projected 2D joint locations do not always align with the PoseTrack21 keypoint annotations. This misalignment stems from the fundamental structural differences between the projected SMPL 2D keypoints and the PoseTrack21 annotations. As a result, visually correct and anatomically consistent mesh reconstructions still yield lower PCK scores. In particular, the pelvis region (e.g., PoseTrack21 keypoints 9 and 12) and the head area (e.g., keypoint 38) highlight these discrepancies. Although our SMPL meshes accurately capture body shape and pose, the corresponding projected keypoints do not coincide with PoseTrack21’s annotation style, leading to systematic offsets in PCK evaluation.

\parahead{Memory Slot Length and Propagation Location Ablation}
In \ours, the memory bank stores a fixed number of historical tracking-query features. Each entry in this memory bank corresponds to a \emph{memory slot}, representing the tracking-query embeddings from previous frames. The \emph{memory slot length} therefore determines the temporal window over which past identity information is retained and made available for aggregation. We conduct an ablation study to analyze how the memory slot length and the (in-model) location at which memory features are propagated affect tracking performance. As shown in \autoref{tab:ablation_trackq_video}, using a shorter memory length (4 slots) consistently yields better performance than using 8 slots.

In \ours, tracking queries are aggregated in the memory bank and then injected into the \ours at different stages. In the \emph{early} strategy, memory is fused before the \ours module, allowing temporal cues to shape the evolution of query embeddings throughout the \ours{} module. 
In contrast, in the \emph{late} strategy, memory is injected after the \ours{} module but still before the prediction heads, allowing temporal cues to refine only the final query representations without affecting the earlier decoding stages. 
While both strategies perform similarly, late propagation slightly improves MOTA and IDF1 on PoseTrack21, suggesting that memory can be more effective when applied after initial query refinement. However, the early propagation variant provides more stable performance across both benchmarks, particularly on BEDLAM where it achieves lower IDs and better MPJPE. For this reason, we adopt the early integration strategy as our default configuration.

Overall, these results show that (1) shorter memory windows are more reliable for multi-person tracking under fast motion, and (2) late propagation yields small gains on PoseTrack21 but early propagation offers a better overall balance across 2D and 3D benchmarks.

\begin{table}[t]
\caption{
\textbf{Ablation study of memory bank parameters.}
Using a shorter \emph{memory slot length} and an \emph{early} injection of memory bank derived features results in the most balanced performance across the PoseTrack21 and BEDLAM datasets and their 2D and 3D metrics.
}
\centering
\resizebox{\columnwidth}{!}{
\setlength{\tabcolsep}{6pt}
\renewcommand{\arraystretch}{1.2}
\begin{tabular}{c c | c c | c c c c}
\toprule
\multirow{2}{*}{Mem Slot} & \multirow{2}{*}{Mem Loc} &
\multicolumn{2}{c|}{PoseTrack21} & 
\multicolumn{4}{c}{BEDLAM} \\
 &  & MOTA$\uparrow$ & IDF1$\uparrow$ &
 MOTA$\uparrow$ & IDF1$\uparrow$ & IDs$\downarrow$ & MPJPE$\downarrow$ \\
\midrule
4 & early & \underline{73.6} & \underline{80.6} & \textbf{99.18} & \textbf{99.37} & \underline{37} &\textbf{48.68} \\
4 & late & \textbf{74.2} & \textbf{81.1} & \underline{99.10} & \underline{99.35} & \textbf{36} & \underline{49.40} \\
8 & early & 73.0 & 79.4 & 98.73 & 99.17 & 61 & 49.82 \\
8 & late & 73.0 & 79.4 & 98.74 & 99.16 & 61 & 49.80 \\
\bottomrule
\end{tabular}
}
\label{tab:ablation_trackq_video}
\end{table}

\parahead{Memory Aggregation Method Ablation} 
%
%
\begin{table}[t]
\caption{
\textbf{Ablation of memory aggregation method.}
Using a similarity-based method of aggregating memory slots and current frame queries out-performs a cross-attention-based alternative.
}
\centering
\resizebox{\columnwidth}{!}{
\setlength{\tabcolsep}{6pt}
\renewcommand{\arraystretch}{1.2}
\begin{tabular}{c | c c | c c c c}
\toprule
\multirow{2}{*}{Mem Aggr.. Med.} &
\multicolumn{2}{c|}{PoseTrack21} & 
\multicolumn{4}{c}{BEDLAM} \\
  & MOTA$\uparrow$ & IDF1$\uparrow$ &
 MOTA$\uparrow$ & IDF1$\uparrow$ & IDs$\downarrow$ & MPJPE$\downarrow$ \\
\midrule
Similarity (Ours) & \textbf{73.6} & \textbf{80.6} & \textbf{99.18} & \textbf{99.37} & \textbf{37} & 48.68 \\
Cross-Attn & 72.2 & 80.2 & 99.15 & 99.34 & \textbf{37} & \textbf{47.68} \\
\bottomrule
\end{tabular}
}
\label{tab:ablation_aggre_method}
\end{table}
We evaluate two different memory propagation mechanisms. Our default approach is the 
\emph{similarity-based aggregation} mechanism, which computes affinities between current 
queries and the stored memory slots as described in the main paper. For comparison, we 
also test a \emph{cross-attention–based} propagation variant that directly attends to the 
stored memory features. As shown in \autoref{tab:ablation_aggre_method}, the similarity-based 
method yields better performance on PoseTrack21 in MOTA, while achieving comparable identity 
preservation on BEDLAM. The cross-attention variant performs similarly overall but offers 
slightly improved 3D accuracy (MPJPE) on BEDLAM. Given this trade-off, the similarity-based mechanism provides a better balance across datasets and remains our default choice.



\section{Identity Embedding Visualization}

\begin{figure}[t]
  \centering
  \resizebox{\columnwidth}{!}{
    \begin{minipage}{\linewidth}
      \centering
      \begin{minipage}{0.32\linewidth}
        \centering
        \includegraphics[width=\linewidth]{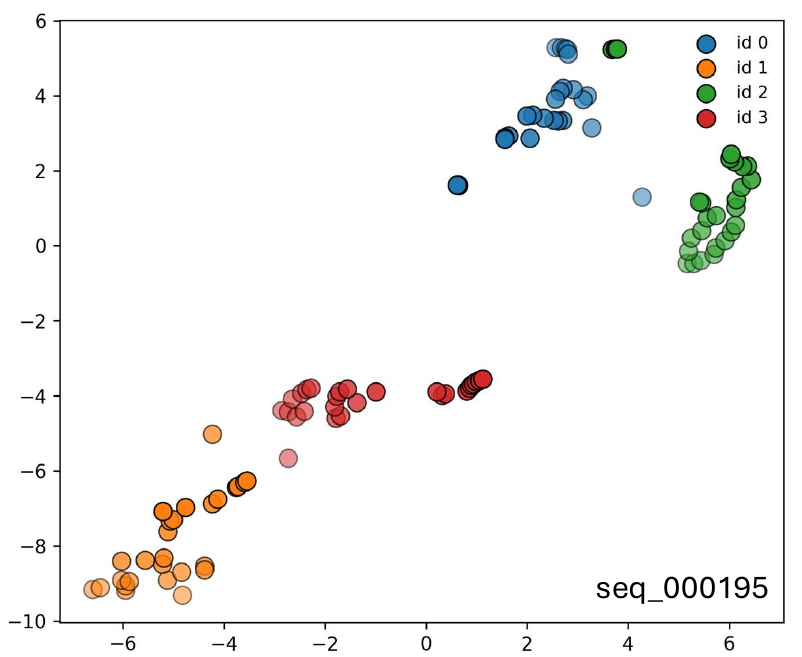}
      \end{minipage}
      \hfill
      \centering
      \begin{minipage}{0.32\linewidth}
        \centering
        \includegraphics[width=\linewidth]{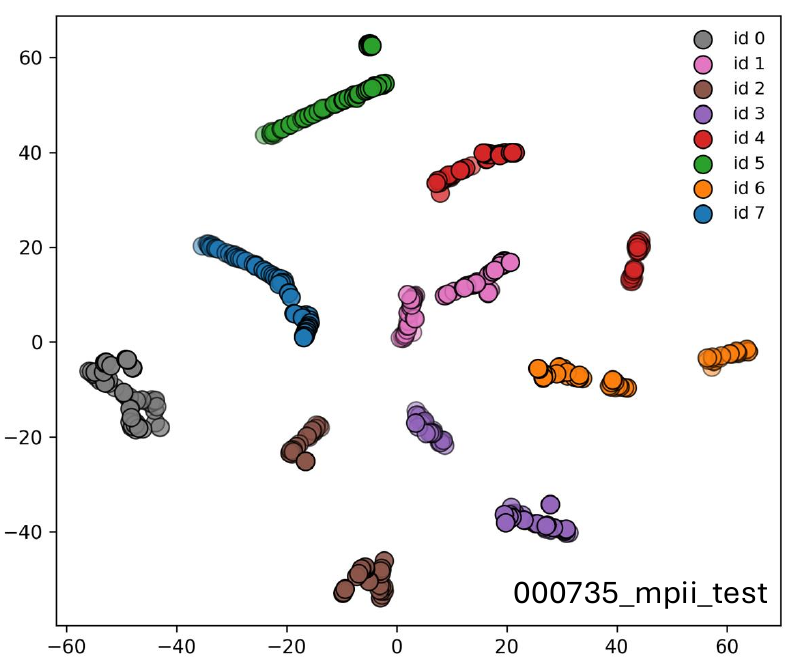}
      \end{minipage}
      \hfill
      \begin{minipage}{0.32\linewidth}
        \centering
        \includegraphics[width=\linewidth]{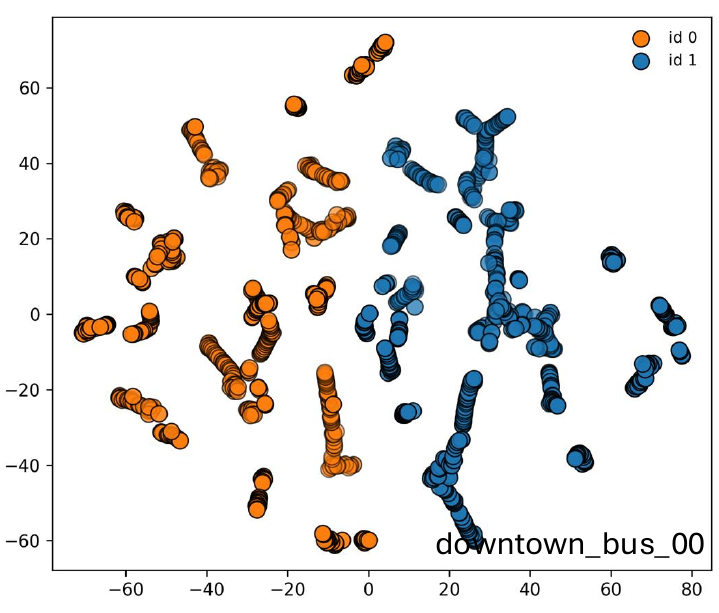}
      \end{minipage}
    \end{minipage}
  }
\caption{t-SNE visualization~\cite{van2008visualizing} of query embeddings over time (darker indicates later frames): BEDLAM (left), PoseTrack (middle), and Dyna3DPW (right).}

  \label{fig:tsne_minipage}
\end{figure}

During inference, each active track is a persistent tracking-query embedding propagated frame-to-frame and updated online via Track-Modulated Cross-Attention (plus retrieved memory if used).
Identity stability is learned through instance-consistent supervision during training (also reflected in IDF1/IDs).
We further visualize this effect using t-SNE~\cite{van2008visualizing} by showing that tracking-query embeddings cluster by ground-truth identity over time (darker points correspond to later frames), with clean separation between identity-specific clusters in the query-embedding space (\autoref{fig:tsne_minipage}).

\section{Further Discussions}

\parahead{Limitations and Future Work}
Our current system supports interaction through user prompts, primarily in the form of bounding boxes. Extending the framework to text-based or high-level semantic instructions remains an 
exciting direction for future work. 
Although our model reconstructs SMPL meshes, it does not recover texture, which limits its ability to capture fine-grained appearance details such as clothes. Incorporating texture or clothing-aware modeling would further enhance realism and broaden the applicability of our system.
Furthermore, the current design focuses on the SMPL body model, and extending the approach to the more expressive SMPL-X representation would improve compatibility with hands, facial expressions, and full-body interaction scenarios.

\parahead{Potential Negative Societal Impacts}
\ours enables analysis of multiple humans in video with the possibility of control (specifying who to track), which can be useful in settings such as sports analytics, human–computer interaction, and safety monitoring. At the same time, any method that detects, tracks, and reconstructs multiple humans over time raises concerns around privacy and potential misuse for non-consensual monitoring. In particular, long-term identity persistence and detailed 3D body reconstructions could be combined with other systems for profiling or re-identification. As with prior work in multi-person tracking and human mesh recovery, responsible use requires appropriate safeguards and considerations on data collection, processing, as well as understanding of dataset and model biases that may lead to uneven performance across demographic groups.
\end{document}